\def\year{2022}\relax
\documentclass[letterpaper]{article} 
\usepackage{aaai22}  
\usepackage{times}  
\usepackage{helvet}  
\usepackage{courier}  
\usepackage[hyphens]{url}  
\usepackage{graphicx} 
\usepackage{natbib}  
\usepackage{caption} 
\DeclareCaptionStyle{ruled}{labelfont=normalfont,labelsep=colon,strut=off} 
\usepackage{algorithm}
\usepackage{algorithmic}

\usepackage{newfloat}
\usepackage{listings}
\floatstyle{ruled}
\newfloat{listing}{tb}{lst}{}
\floatname{listing}{Listing}

\usepackage[utf8]{inputenc}
\usepackage{adjustbox}
\usepackage{amsmath}
\usepackage{amssymb}
\usepackage[title,titletoc,page]{appendix}
\usepackage{bbm}

\usepackage{booktabs}
\usepackage{tabularx}
\usepackage{float}
\usepackage{placeins}
\usepackage{ifthen}
\usepackage{tikz}
\usepackage{colortbl}
\usepackage{caption}
\usepackage{subcaption}
\usepackage{blindtext}
\usepackage{environ}
\usepackage{multirow}
\usepackage{stmaryrd}

\usetikzlibrary{shapes, arrows, positioning}
\usetikzlibrary{decorations.pathreplacing}

\DeclareMathOperator*{\argmax}{arg\,max}

\DeclareMathOperator{\softmax}{softmax}

\nocopyright
\title{SEA: Graph Shell Attention in Graph Neural Networks}
\author{
    Christian M.M. Frey, Yunpu Ma, Matthias Schubert
}
\affiliations{
    Institute for Informatics

    Oettingenstr. 67\\
    80538 Munich, Germany\\
    Christian.Frey@lmu.de, \{ma, schubert\}@dbs.ifi.lmu.de
}

\usepackage{bibentry}

\begin{document}

\maketitle

\begin{abstract}
A common issue in \emph{Graph Neural Networks} (GNNs) is known as \emph{over-smoothing}. By increasing the number of iterations within the message-passing of GNNs, the nodes' representations of the input graph align with each other and become indiscernible.
Recently, it has been shown that increasing a model's complexity by integrating an attention mechanism yields more expressive architectures. This is majorly contributed to steering the nodes' representations only towards nodes that are more informative than others. Transformer models in combination with GNNs result in architectures including \emph{Graph Transformer Layers} (GTL), where layers are entirely based on the attention operation. 
However, the calculation of a node's representation is still restricted to the computational working flow of a GNN. 
In our work, we relax the GNN architecture by means of implementing a routing heuristic. Specifically, the nodes' representations are routed to dedicated experts. Each expert calculates the representations according to their respective GNN workflow. The definitions of distinguishable GNNs result from $k$-localized views starting from the central node. We call this procedure \emph{Graph \textbf{S}h\textbf{e}ll \textbf{A}ttention} (SEA), where experts process different subgraphs in a transformer-motivated fashion.
Intuitively, by increasing the number of experts, the models gain in expressiveness such that a node's representation is solely based on nodes that are located within the receptive field of an expert. 
We evaluate our architecture on various benchmark datasets showing competitive results compared to state-of-the-art models.
\end{abstract}





\section{Introduction}
\label{sec:sea_introduction}
The modeling flow of Graph Neural Networks (GNNs) has been proven to be a convenient tool in a variety of real-world applications building on top of graph data \cite{wu_2021_survey_gnn}. These range from predictions in social networks over property predictions in molecular graph structures to content recommendations in online platforms. From a machine learning perspective, we can categorize them into various theoretical problems that are known as \emph{node classification}, \emph{graph classification/regression} - encompassing binary decisions or modeling a continuous-valued function -, and \emph{relation prediction}. In our work, we propose a novel framework and show its applicability on graph-level classification and regression, as well as on node-level classification tasks.

The high-level intuition behind GNNs is that by increasing the number of iterations $k=1,\ldots, K$, a node's representation processes, contains, and therefore relies more and more on its $k$-hop neighborhood. However, a well-known issue with the vanilla GNN architecture refers to a problem called \emph{over-smoothing} \cite{hoory_2006_expandergraphs, xu_2018_howpowerfulgnn}. 
Suppose we are given a GNN-motivated model, the information flow between two nodes $u,v \in \mathcal{V}$, where $\mathcal{V}$ denotes a set of nodes, is proportional to the reachability of node $v$ on a $k$-step random walk starting from $u$. Hence, by increasing the layers within the GNN architecture, the information flow of every node approaches the stationary distribution of random walks over the graph. As a consequence, the localized information flow is getting lost. 
On graph data that follow strong connectivity, it takes $k = O(log |\mathcal{V}|)$ steps for a random walk starting from an arbitrary node to converge to an (almost) uniform distribution. Consequently, increasing the number of iterations within the GNN message-passing results in representations for all the nodes in the input graph that align and become indiscernible.

One strategy for increasing a GNN's expressiveness is by adding an attention mechanism into the architecture. Recently, it has been shown how the \emph{Transformer} model \cite{Vaswani_attention_2017} can be applied on graph data \cite{dwivedi_2021_generalization} yielding competitive results to state-of-the-art models. Generally, multi-headed attention shows vying results whenever we have prior knowledge to indicate that some neighbors might be more informative than others. 

Our framework further improves the representational capacity by adding an expert motivated heuristic into the GNN architecture. More specifically, to compute a node's representation, a routing module first decides upon an expert that is responsible for a node's computation. The experts differ in how their $k$-hop localized neighborhood is processed and capture various depths of GNNs. We refer to the different substructures that individual experts process by \emph{Graph Shells}. As each expert attends to a specific subgraph of the input graph, we introduce the concept of \emph{Graph \textbf{S}h\textbf{e}ll \textbf{A}ttention} (SEA). Hence, whereas a vanilla GNN lacks in over-smoothing the representations for all nodes, we introduce additional degrees of freedom in our architecture to simultaneously capture short- and long-term dependencies being processed by respective experts.\hfill\\

\noindent%
In summary, our contributions are as follows:
\begin{itemize}
    \item Integration of expert-routing mechanism into Transformer motivated models applied on graph data;
    \item Novel Graph Shell Attention (SEA) models relaxing the vanilla GNN definition;
\end{itemize}

\section{Related Work}
\label{sec:sea_relatedWork}
In recent years, the AI community proposed various forms of (self-)attention in numerous domains. \emph{Attention} itself refers to a mechanism in neural networks where a model learns to make predictions by selectively attending to a given set of data. The success of these models dates back to Vaswani et al. \cite{Vaswani_attention_2017} by introducing the \emph{Transformer} model. It relies on scaled dot-product attention, i.e., given a query matrix $Q$, a key matrix $K$, and a value matrix $V$, the output is a weighted sum of the value vectors, where the dot-product of the query with corresponding keys determines the weight that is assigned to each value. 

Transformer architectures have also been successfully applied to graph data. A thorough work by Dwivedi et al. \cite{dwivedi_2021_generalization} evaluates transformer-based GNNs. They conclude that the attention mechanism in Transformers applied on graph data should only aggregate the information from the local neighborhood, ensuring graph sparsity. As in \emph{Natural Language Processing} (NLP), where a positional encoding is applied, they propose to use Laplacian eigenvectors as the positional encodings for further improvements. In their results, they outperform baseline GNNs on the graph representation task. A similar work \cite{kreuzer_2021_rethinking} proposes a full Laplacian spectrum to learn the position of each node within a graph. 
Yun et al. \cite{yun_2019_GTN} proposed \emph{Graph Transformer Networks} (GTN) that is capable of learning on heterogeneous graphs. The target is to transform a given heterogeneous input graph into a meta-path-based graph and apply a convolution operation afterward. Hence, the focus of their attention framework is on interpreting generated meta-paths.
Another transformer-based architecture that has been introduced by Hu et al. \cite{hu_2020_hgt} is \emph{Heterogeneous Graph Transformer} (HGT). Notably, their architecture can capture graph dynamics w.r.t the information flow in heterogeneous graphs. Specifically, they take the relative temporal positional encoding into account based on differences of temporal information given for the central node and the message-passing nodes. 
By including the temporal information, Zhou et al. \cite{zhou_2020_temporalinteraction} built a transformer-based generative model for generating temporal graphs by directly learning from the dynamic information in networks. 
The work of Ngyuen et al. \cite{Nguyen_2019_UGT} proposes another idea for positional encoding.
The authors of this work introduced a graph transformer for arbitrary homogeneous graphs with a coordinate embedding-based positional encoding scheme. 

In \cite{ying2021_graphormer}, the authors introduced a transformer motivated architecture where various encodings are aggregated to compute the hidden representations. They propose graph structural encodings subsuming a spatial encoding, an edge encoding, and a centrality encoding. 

Furthermore, a work exploring the effectiveness of large-scale pre-trained GNN models is proposed by the \emph{GROVER} model \cite{Rong_2020_selfsupervisedGraphTransformer}. The authors include an additional GNN applied in the attention sublayer to produce vectors for $Q$, $K$, and $V$. Moreover, they apply single long-range residual connections and two branches of feedforward networks to produce node and edge representations separately. In a self-supervised fashion, they first pre-train their model on $10$ million unlabeled molecules before using the resulting node representations in downstream tasks.

Typically, all the models are built in a way such that the same parameters are used for all inputs. To gain more expressiveness, the motivation of the mixture of experts (MoE) heuristic \cite{shazeer_2017_moelayer} is to apply different parameters w.r.t to the input data. Recently, Google proposed \emph{Switch Transformer} \cite{fedus_2021_switchTransformer}, enabling training above a trillion parameter networks but keeping the computational cost in the inference step constant. In our work, we will show how we can apply MoE motivated heuristics in the scope of GNN models.

\section{Preliminaries}
\label{sec:sea_preliminaries}
In this section, we provide definitions and recap on the general \emph{message-passing} paradigm combined with \emph{Graph Transformer Layers} (GTLs) \cite{dwivedi_2021_generalization}.

\subsection{Notation}
Let $\mathcal{G}=(\mathcal{V}, \mathcal{E})$ be an undirected graph where $\mathcal{V}$ denotes a set of nodes and $\mathcal{E}$ denotes a set of edges connecting nodes. We define $\mathcal{N}_k(u)$ to be the $k$-hop neighborhood of a node $u \in \mathcal{V}$, i.e., 
$N_k(u)=\{v \in \mathcal{V}: d_\mathcal{G}(u,v) \leq k\}$, where $d_G(u,v)$ denotes the hop-distance between $u$ and $v$ on $\mathcal{G}$. For $N_1(u)$ we will simply write $N(u)$ and omit the index $k$. The induced subgraph by including the $k$-hop neighbors starting from node $u$ is denoted by $\mathcal{G}_u^k$. Moreover, in the following we will use a real-valued representation vector $h_u \in \mathbb{R}^{d}$ for a node $u$, where $d$ denotes the embedding dimensionality.

\subsection{Graph Neural Networks}
\label{subsec:sea_gnn}
Given a graph $\mathcal{G}=(\mathcal{V}, \mathcal{E})$ with node attributes $X_{\mathcal{V}}=\{X_{u} | u \in \mathcal{V}\}$ and edge attributes $X_\mathcal{E} = \{X_{uv} |(u,v) \in \mathcal{E}\}$, a GNN aims to learn an embedding vector $h_u$ for each node $u \in \mathcal{V}$, and a vector $h_\mathcal{G}$ for the entire graph $\mathcal{G}$. For an $L$-layer GNN, a neighborhood aggregation scheme is performed to capture the $L$-hop information surrounding each node. The $l$-th layer of a GNN is formalized as follows:

\begin{align}
    h_u^{l+1} &= \textsc{update}^{l} (h_u^{l}, m_{N(u)}^{l}) \label{eq:sea_gnn_update}\\ 
    m_{N(u)}^{l} &= \textsc{aggregate}^{l}(\{(h_v^{l}) : v \in N(u)\}),\label{eq:sea_gnn_aggregate}
\end{align}

\noindent%
where $N(u)$ is the $1$-hop neighborhood set of $u$, $h_u^{(l)}$ denotes the representation of node $u$ at the $l$-th layer, and $h_u^{(0)}$ is initialized as the node attribute $X_u$. Since $h_u$ summarizes the information of central node $u$, it is also referred as $\textit{patch embedding}$ in the literature. A graph's embedding $h_\mathcal{G}$ is derived by a permutation-invariant readout function:
\begin{equation}
    h_\mathcal{G} = \textsc{readout}(\{h_u | u \in \mathcal{V}\})
\end{equation}

A common heuristic for the readout function is to choose a function $\textsc{readout}(\cdot) \in \{mean(\cdot), sum(\cdot), max(\cdot)\}$.

\subsection{Transformer}
\label{subsec:sea_transformer}
The vanilla Transformer architecture as proposed by Vaswani et al. \cite{Vaswani_attention_2017} was originally introduced in the  scope of \emph{Natural Language Processing} (NLP) and consists of a composition of \emph{Transformer layers}. Each Transformer layer has two parts: a self-attention module and a position-wise feedforward network (FFN). Let $H=[h_1^T, \ldots, h_n^T]^T \in \mathbb{R}^{n \times d}$ denote the input of the self-attention module where $d$ is the hidden dimension and $h_i \in \mathbb{R}^{1 \times d}$ is the hidden representation at position $i$ of an input sequence. The input $H$ is projected by three matrices $W_Q \in \mathbb{R}^{d \times d_Q}$, $W_K \in \mathbb{R}^{d \times d_K}$, and $W_V \in \mathbb{R}^{d \times d_V}$ to get the corresponding representation $Q, K, V$. The self-attention is calculated as: 
\begin{equation}
\begin{split}
    Q = HW_Q \qquad K = HW_K \qquad V = HW_V,\\
    A = \frac{QK^T}{\sqrt{d_k}}, \quad Attn(H) = \softmax(A)V    
\end{split}
\end{equation}
Notably, in NLP as well as in computer vision tasks, the usage of transformer models were boosters behind a large number of state-of-the-art systems. Recently, the Transformer architecture has also been modified to be applicable to graph data \cite{dwivedi_2021_generalization}, which we will briefly recap in the next section. 

\subsection{Graph Transformer Layer}
\label{subsec:sea_gtl}

\begin{figure}
\resizebox{0.95\linewidth}{!}{
\tikzset{%
  outerblock/.style    = {draw, dotted, thick, rectangle, minimum height = 5cm,
    minimum width = 9cm},
  innerblock/.style    = {draw, thick, rectangle, minimum height = 3em,
    minimum width = 3em},
  sum/.style      = {draw, circle, node distance = 2cm}, 
  input/.style    = {coordinate}, 
  output/.style   = {coordinate}, 
  textblock/.style={rectangle,draw, fill=gray!10, minimum width=3.25cm, minimum height=0.5cm, rounded corners=0.8ex, anchor=center, inner sep=0},
  textblock_red/.style={rectangle,draw,fill=red!20,rounded corners=.8ex},
  summation/.style={path picture={\draw[black]
    (path picture bounding box.south) -- (path picture bounding box.north) (path picture bounding box.west) -- (path picture bounding box.east);}},
}

\definecolor{lavander}{cmyk}{0,0.48,0,0}
\definecolor{violet}{cmyk}{0.79,0.88,0,0}
\definecolor{burntorange}{cmyk}{0,0.52,1,0}
\definecolor{blue}{cmyk}{0.8,0.4,0.4,0}
\definecolor{green}{cmyk}{0.8, 0.0, 0.6, 0}

\def\lav{lavander!90}
\def\oran{orange!30}
\def\blue{blue!30}
\def\green{green!30}

\tikzstyle{one_hop}=[draw, circle, 
violet, 
bottom color=\lav, 
top color= white, 
text=violet, minimum width=10pt]
\tikzstyle{superpeers}=[draw, circle, burntorange, left color=\oran, text=violet, minimum width=10pt]
\tikzstyle{two_hop}=[draw, circle, blue, bottom color=\blue, top color= white, text=violet, minimum width=10pt]

\tikzstyle{legend_general}=[rectangle, rounded corners, thin, burntorange, fill= white, draw, text=violet, minimum width=2.5cm, minimum height=0.8cm]
\tikzstyle{legend_general_pos}=[rectangle, rounded corners, thin, green, fill= white, draw, text=violet, minimum width=2.5cm, minimum height=0.8cm]
\tikzstyle{legend_general_neg}=[rectangle, rounded corners, thin, red, fill= white, draw, text=violet, minimum width=2.5cm, minimum height=0.8cm]
\tikzstyle{module}=[rectangle, rounded corners, thin, blue, fill= white, draw, text=blue, minimum width=2.0cm, minimum height=0.8cm]
\tikzstyle{module_neg}=[rectangle, rounded corners, thin, red, fill= white, draw, text=blue, minimum width=2.0cm, minimum height=0.8cm]
                           
\begin{tikzpicture}[auto, thick, node distance=0.75cm]

\begin{scope}
\node[draw, fill=orange!10, dotted, rectangle, minimum width=13.5cm, minimum height=12cm, rounded corners=1ex] at (4.75, 5) {};
\node[draw, fill=blue!10, opacity=0.4,dotted, rectangle, minimum width=11.5cm, minimum height=6.5cm, rounded corners=1ex] at (4.75,2.75) {};

\node[textblock, minimum width=1cm, fill=gray!20] (hi) at (0,-2.5) {$h_u$};
\node[circle, draw, summation] (lap_hi) at (0, -1.5) {};
\node[circle, draw] (lambda_hi) at (1, -1.5) {$\lambda$};

\node[textblock, minimum width=1cm, fill=gray!20] (hj) at (4.5,-2.5) {$\{h_v\}$};
\node[circle, draw, summation] (lap_hj) at (4.5, -1.5) {};
\node[circle, draw] (lambda_hj) at (5.5, -1.5) {$\lambda$};

\node[textblock, minimum width=1cm, fill=gray!20] (eij) at (9,-1.5) {$\{e_{uv}\}$};

\node[textblock, minimum width=1cm, fill=cyan!20] (Q) at (0,0) {$Q$};
\node[textblock, minimum width=1cm, fill=cyan!20] (K) at (3,0) {$K$};
\node[textblock, minimum width=1cm, fill=cyan!20] (V) at (6,0) {$V$};
\node[textblock, minimum width=1cm, fill=cyan!20] (E) at (9,0) {$E$};
\node[textblock] (product) at (1.5,1.5) {Product};
\node[textblock] (scaling) at (1.5,2.5) {Scaling};
\node[circle, draw] (dot_scale) at (1.5, 3.25) {$\bullet$};
\node[textblock, fill=green!20] (softmax) at (1.5,4.5) {softmax};
\node[circle, draw] (dot) at (6, 4.5) {$\bullet$};
\node[textblock] (sum) at (6,5.5) {sum};
\node[textblock] (concat) at (3,6.5) {Concat};
\node[textblock, fill=cyan!20] (linear1) at (3,7.5) {Linear};
\node[textblock, fill=red!20] (addnorm1) at (3,8.5) {Add \& Normalization};
\node[textblock, fill=cyan!20] (linear2) at (3,9.5) {Linear};
\node[textblock, fill=red!20] (addnorm2) at (3,10.5) {Add \& Normalization};

\node[textblock] (concat_edge) at (9,6.5) {Concat};
\node[textblock, fill=cyan!20] (linear1_edge) at (9,7.5) {Linear};
\node[textblock, fill=red!20] (addnorm1_edge) at (9,8.5) {Add \& Normalization};
\node[textblock, fill=cyan!20] (linear2_edge) at (9,9.5) {Linear};
\node[textblock, fill=red!20] (addnorm2_edge) at (9,10.5) {Add \& Normalization};

\draw[->] (Q) -- (0, 0.75) -- (1, 0.75) -- (product);
\draw[->] (K) -- (3, 0.75) -- (2, 0.75) -- (product);
\draw[->] (product) -- (scaling);
\draw[->] (scaling) -- (dot_scale);
\draw[->] (dot_scale) -- (softmax);
\draw[->] (softmax) -- (dot);
\draw[->] (dot) -- (sum);
\draw[->] (sum) -- (3,5.5) -- (concat);
\draw[->] (concat) -- (linear1);
\draw[->] (linear1) -- (addnorm1);
\draw[->] (addnorm1) -- (linear2);
\draw[->] (linear2) -- (addnorm2);

\draw[->] (concat_edge) -- (linear1_edge);
\draw[->] (linear1_edge) -- (addnorm1_edge);
\draw[->] (addnorm1_edge) -- (linear2_edge);
\draw[->] (linear2_edge) -- (addnorm2_edge);
\draw[->] (E) -- (9,3.25) -- (dot_scale);
\draw[->] (1.5, 3.75) -- (9,3.75) -- (concat_edge);

\draw[->] (V) -- (dot);

\draw[->] (hi) -- (lap_hi);
\draw[->] (lap_hi) -- (Q);

\draw[->] (hj) -- (lap_hj);
\draw[->] (lap_hj) -- (4.5, -0.75) -- (3, -0.75) -- (K);
\draw[->] (lambda_hi) -- (lap_hi);
\draw[->] (lap_hj) -- (4.5, -0.75) -- (6, -0.75) -- (V);
\draw[->] (lambda_hj) -- (lap_hj);
\draw[->] (eij) -- (E);

\draw[->] (0, -0.75) -- (-1.5, -0.75) -- (-1.5, 8.5) -- (addnorm1);
\draw[->] (3, 9) -- (-1.5, 9) -- (-1.5, 10.5) -- (addnorm2);

\draw[->] (9, -0.75) -- (11, -0.75) -- (11, 8.5) -- (addnorm1_edge);
\draw[->] (9, 9) -- (11, 9) -- (11, 10.5) -- (addnorm2_edge);

\end{scope}

\end{tikzpicture}
}
\caption{Vanilla Graph Transformer Layer with edge features \cite{dwivedi_2021_generalization}; $\lambda$ denotes the Laplacian eigenvectors for positional encodings} 
\label{fig:sea_gtl_with_edges}
\end{figure}
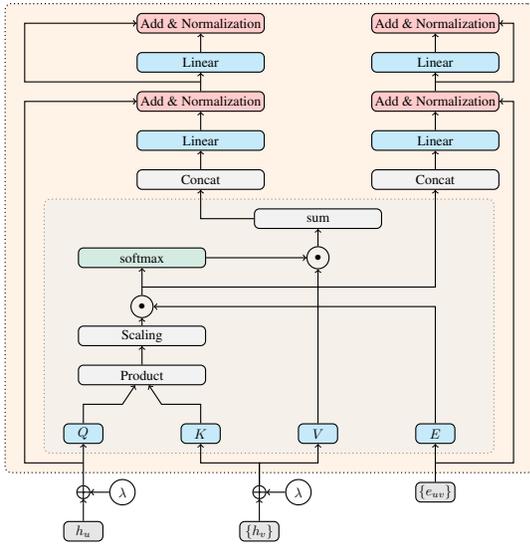

Generally, in NLP, the words in an input sequence can be represented as a fully connected graph that includes all connections between the input words. However, such an architecture does not leverage graph connectivity. Therefore, it can perform poorly whenever the graph topology is important and has not been encoded in the input data in any other way. 
These thoughts lead to the work \cite{dwivedi_2021_generalization}, where a general \emph{Graph Transformer Layer} (GTL) has been introduced. 

For the sake of completeness, we will recap on the most important equations. A layer update for layer $l$ within a GTL including edge features is defined as:

\begin{align}
    \hat{h}_u^{l+1} &= O_h^l \bigparallel_{k=1}^H (\sum_{v \in \mathcal{N}_u} w_{uv}^{k,l}V^{k,l}h_v^l),\\
    \hat{e}_{uv}^{l+1} &= O_e^l \bigparallel_{k=1}^H (\hat{w}_{uv}^{k,l}), \text{where},\\
    w_{uv}^{k,l} &= \softmax_v(\hat{w}_{uv}^{k,l}),\\
    \hat{w}_{uv}^{k,l} &= (\frac{Q^{k,l}h_u^l \cdot K^{k,l}h_v^l}{\sqrt{d_k}})\cdot E^{k,l}e_{uv}^l,
\end{align}
where $Q^{k,l}, K^{k,l}, V^{k,l}, E^{k,l} \in \mathbb{R}^{d_k \times d}$, and $O_h^l,O_e^l \in \mathbb{R}^{d \times d}$. The operator $\bigparallel$ denotes the concatenation of attention heads $k=1, \ldots, H$. Subsequently, the outputs $\hat{h}_u^{l+1}$ and $\hat{e}_{uv}^{l+1}$ are passed to feedforward networks and succeeded by residual connections and normalization layers as in the vanilla Transformer architecture \cite{Vaswani_attention_2017}. The full architecture of GTLs is illustrated in figure \ref{fig:sea_gtl_with_edges}.\hfill\\

\noindent\textbf{Positional Encoding. } Using the Laplacian eigenvectors as the positional encoding for nodes within a graph is a common heuristic \cite{dwivedi_2021_generalization}. From spectral analysis, the eigenvectors are computed by the factorization of the Laplacian matrix: 
\begin{equation}
    \Delta = I - D^{-1/2}AD^{-1/2} = U^T \Lambda U,
\end{equation}
where $A$ denotes the $n \times n$ adjacency matrix, $D$ denotes the degree matrix. The decomposition yields $\Lambda, U$ corresponding to the eigenvalues and eigenvectors. For the positional encoding, we use the $k$ smallest eigenvectors of a node. 
In our models, we also apply the \emph{Laplacian Positional Encoding} (LPE).
\section{Methodology}
\label{sec:sea_methodology}
In this section, we introduce our \emph{Graph \textbf{S}h\textbf{e}ll \textbf{A}ttention} (SEA) architecture for graph data. SEA builds on top of the message-passing paradigm of \emph{Graph Neural Networks} (GNNs) and integrates an expert heuristic.

\emph{Over-smoothing} in GNNs is a well-known issue \cite{xu_2018_howpowerfulgnn} and exacerbates the problem when we build deeper GNN models which generate similar representations for all nodes in the graph. Applying the same number of iterations for each node inhibits the expressiveness of short- and long-term dependencies simultaneously. Therefore, our architecture loosens up the strict workflow that is applied for each node equally.
We gain expressiveness by routing each node representation towards dedicated experts, which only process nodes in their $k$-localized receptive field.


\subsection{Graph Shells}
\label{subsec:sea_graphShells}
In our approach, we exploit the graph transformer architecture, including \emph{Graph Transformer Layers} (GTLs) \cite{dwivedi_2021_generalization} and extend it by a set of \emph{experts}. A routing layer decides upon which expert is most relevant for the computation of a node's representation. Intuitively, the expert's computation for a node representation differs in how $k$-hop neighbors are stored and processed within GTLs. 


Generally, starting from a central node, \emph{Graph Shells} refer to subgraphs that include only nodes that have at maximum a $k$-hop distance ($k$-neighborhood). 
More formally, the $i$-th expert comprises the information given in the $i$-th neighborhood $N_{i}(u)=\{v \in \mathcal{V}: d_G(u,v) \leq i)\}$, where $u \in \mathcal{V}$ denotes the central node. We refer to the subgraph $\mathcal{G}_u^{i}$ as the expert's \emph{receptive field}.  
Notably, increasing the number of iterations within GNNs correlate with the number of experts being used. 

In the following, we introduce three variants on how experts process graph shells. 


\subsubsection{SEA-\textsc{gnn}}
\label{subsubsec:sea_sea_gtl}\hfill\\
For the definition of the first graph shell model, we exploit the GNN architecture. Hence, the shells described by each expert are given by construction. From the formal definition expressed in eqs. \ref{eq:sea_gnn_update} and \ref{eq:sea_gnn_aggregate}, the information for the $l$-th expert is defined by the $l$-th iteration of the GNN. For $N$ experts, we set the maximal number of hops to be $L=N$. Figure \ref{fig:sea_graphShell_gnnExpert} illustrates this graph shell model. From left to right, the information of nodes being reachable by more hops is processed. Experts storing information after few hops refer to short-term dependencies, whereas experts processing more hops yields information of long-term dependencies. 

\subsubsection{SEA-\textsc{aggregated}}
\label{subsubsec:sea_sea_aggregated}\hfill\\
For the computation of the hidden representation $h_u^{l+1}$ for node $u$ on layer $l+1$, the second model employs an aggregated value from the previous iteration. According to the message-passing paradigm of a vanilla GNN, the aggregation function of eq. \ref{eq:sea_gnn_aggregate} considers the $1$-hop neighbors $N_1(u)$. 
For \emph{SEA-\textsc{aggregated}}, we send the aggregated value $m_{N_{1}}(u)^l$ to all its $1$-hop neighbors. For a node $v \in N_1(u)$, the values received by $v$ are processed according to another aggregation function which can be chosen among $\textsc{Aggregate}_\mu \in \{mean(\cdot), sum(\cdot), max(\cdot)\}$. Formally:
\begin{equation}
    h_u^{l+1} = \textsc{Aggregate}_\mu^{l} ((m_{N(v)}^{l}) : v \in N(u)\})
\end{equation}

\noindent%
Figure \ref{fig:sea_graphShell_AggShells} illustrates this graph shell model. In the first iteration, there are no proceeding layers. Hence, the first expert processes in the same way as the first model. The representations for the second expert are computed by taking the aggregated values of the first shell into account. Sending values to neighboring nodes is illustrated by a full-colored shell.  

\subsubsection{SEA-\textsc{k-hop}}
\label{subsubsec:sea_sea_khop}
\begin{figure}
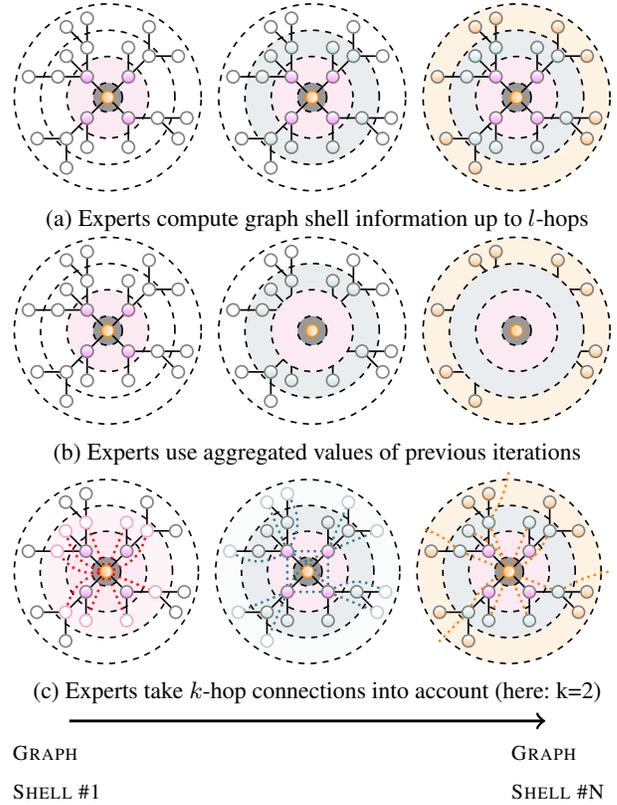

    \centering
    \begin{subfigure}[t]{0.96\linewidth}
        \centering
        \resizebox{1\textwidth}{!}{%
            \input{_tikz_architecture_gnnexpert}
        }
        \caption{Experts compute graph shell information up to $l$-hops}
        \label{fig:sea_graphShell_gnnExpert}
    \end{subfigure}
    \hfill
    \begin{subfigure}[t]{0.96\linewidth}
        \centering
        \resizebox{1\textwidth}{!}{%
            \input{_tikz_architecture_aggShells}
        }
        \caption{Experts use aggregated values of previous iterations}
        \label{fig:sea_graphShell_AggShells}
    \end{subfigure}
    \hfill
    \begin{subfigure}[t]{0.96\linewidth}
        \centering
        \resizebox{1\textwidth}{!}{%
            \input{_tikz_architecture_twohopExpert}
        }
        \caption{Experts take $k$-hop connections into account (here: k=2)}
        \label{fig:sea_graphShell_twohops}
    \end{subfigure}
    \begin{subfigure}[t]{1\linewidth}
        \centering
        \resizebox{1\textwidth}{!}{%
            \begin{tikzpicture}[node distance=0.cm]
\node (A) at (0,0) {};
\node (B) at (5,0) {};
\draw[->, thick] (A) -- (B);
\node[below = of A, align=left] {\tiny\textsc{Graph}\\\tiny\textsc{Shell \#1}}; 
\node[below = of B, align=left] {\tiny\textsc{Graph}\\\tiny\textsc{Shell \#N}}; 
\end{tikzpicture}
        }
    \end{subfigure}
    \caption{Three variants of SEA models; for each model, the respective fields of $3$ experts are shown from left to right}
    \label{fig:sea_graphShell}
\end{figure}

\hfill\\
For this model we relax the \textsc{aggregate} function defined in eq. \ref{eq:sea_gnn_aggregate}. Given a graph $\mathcal{G}$, we also consider $k$-hop linkages in the graph connecting a node $u$ with all entities having a maximum distance of $d_\mathcal{G}(u, v)=k$. The relaxations of eqs. \ref{eq:sea_gnn_update} and \ref{eq:sea_gnn_aggregate} can then be written as:
\begin{align}
        h_u^{l+1} &= \textsc{update}^{l} (h_u^{l}, m_{N_k(u)}^{l}) \label{eq:sea_gnn_update_knn}\\ 
        m_{N_k(u)}^{l} &= \textsc{aggregate}^{l}(\{(h_v^{l}) : v \in N_k(u)\}),\label{eq:sea:gnn_aggregate_knn},
\end{align}
where $N_k(u)$ denotes the $k$-hop neighborhood set. This approach allows for processing each $N_1(u), \ldots, N_k(u)$ with their own submodules, i.e., for each $k$-hop neighbors we use respective feedforward networks to compute $Q, K, V$ in GTLs.
Notably, this definition can be interpreted as a generalization of the vanilla GNN architecture, which is given by setting $k=1$.
Figure \ref{fig:sea_graphShell_twohops} shows the $k$-hop graph shell model with $k=2$. From the first iteration on, we take the $k$-hop neighbors into account to calculate a node's representation. Dotted edges illustrate the additional information flow.

\subsection{SEA: Shell Attention}
\label{subsec:sea_shellAttention}
By endowing our models with experts referring to various graph shells, we gain expressiveness. A routing module decides to which graph shell the attention is steered. We apply a \emph{single expert} strategy \cite{fedus_2021_switchTransformer}. 
\begin{figure}
\resizebox{0.95\linewidth}{!}{
\tikzset{%
  outerblock/.style    = {draw, dotted, thick, rectangle, minimum height = 5cm,
    minimum width = 9cm},
  innerblock/.style    = {draw, thick, rectangle, minimum height = 3em,
    minimum width = 3em},
  sum/.style      = {draw, circle, node distance = 2cm}, 
  input/.style    = {coordinate}, 
  output/.style   = {coordinate}, 
  textblock/.style={rectangle,draw, fill=gray!10, minimum width=3cm, minimum height=0.5cm, rounded corners=0.8ex, anchor=center, inner sep=0},
  textblock_red/.style={rectangle,draw,fill=red!20,rounded corners=.8ex},
}

\definecolor{lavander}{cmyk}{0,0.48,0,0}
\definecolor{violet}{cmyk}{0.79,0.88,0,0}
\definecolor{burntorange}{cmyk}{0,0.52,1,0}
\definecolor{blue}{cmyk}{0.8,0.4,0.4,0}
\definecolor{green}{cmyk}{0.8, 0.0, 0.6, 0}

\def\lav{lavander!90}
\def\oran{orange!30}
\def\blue{blue!30}
\def\green{green!30}

\tikzstyle{one_hop}=[draw, circle, 
violet, 
bottom color=\lav, 
top color= white, 
text=violet, minimum width=10pt]
\tikzstyle{superpeers}=[draw, circle, burntorange, left color=\oran, text=violet, minimum width=10pt]
\tikzstyle{two_hop}=[draw, circle, blue, bottom color=\blue, top color= white, text=violet, minimum width=10pt]

\tikzstyle{legend_general}=[rectangle, rounded corners, thin, burntorange, fill= white, draw, text=violet, minimum width=2.5cm, minimum height=0.8cm]
\tikzstyle{legend_general_pos}=[rectangle, rounded corners, thin, green, fill= white, draw, text=violet, minimum width=2.5cm, minimum height=0.8cm]
\tikzstyle{legend_general_neg}=[rectangle, rounded corners, thin, red, fill= white, draw, text=violet, minimum width=2.5cm, minimum height=0.8cm]
\tikzstyle{module}=[rectangle, rounded corners, thin, blue, fill= white, draw, text=blue, minimum width=2.0cm, minimum height=0.8cm]
\tikzstyle{module_neg}=[rectangle, rounded corners, thin, red, fill= white, draw, text=blue, minimum width=2.0cm, minimum height=0.8cm]
                           
\begin{tikzpicture}[auto, thick, node distance=0.75cm]

\begin{scope}
\node[draw, fill=orange!30, dotted, rectangle, minimum width=7.5cm, minimum height=3.5cm, rounded corners=1ex] at (6.25, 4.00) {};
\node[draw, fill=green!20, dotted, rectangle, minimum width=7.5cm, minimum height=3.5cm, rounded corners=1ex] at (6.00, 4.75) {};

\begin{scope}[transform canvas={scale=0.8}, yshift=6.5cm, xshift=13cm, node distance=0.6cm, every node/.append style={scale=0.8}]
     \filldraw[fill=white!10!white,thick,dashed] (-3, -0.5) circle (1.6cm);    \filldraw[fill=blue!10!white,thick,dashed] (-3, -0.5) circle (1.15cm);
    \filldraw[fill=lavander!20!white,thick,dashed] (-3, -0.5) circle (0.7cm);
    \filldraw[fill=white!60!black,thick,dashed] (-3, -0.5) circle (0.25cm);

    \foreach \place/\name in {{(-3,-0.5)/e}}
        \node[superpeers] (\name) at \place {};
   %
    \foreach \pos/\i in {above left of/1, below right of/2, below left of/3, above right of/4}
        \node[one_hop, \pos=e] (e\i) {};

    \foreach \speer/\peer in {e/e1,e/e2,e/e3,e/e4}
        \path (\speer) edge (\peer);

    \foreach \pos/\i in {above left of/1, left of/2, above of/3}
        \node[two_hop, \pos=e1] (f\i) {};
    \foreach \speer/\peer in {e1/f1, e1/f2, e1/f3}
        \path (\speer) edge (\peer);

    \foreach \pos/\i in {right of/1, below of/2}
        \node[two_hop, \pos=e2] (g\i) {};
    \foreach \speer/\peer in {e2/g1, e2/g2}
        \path (\speer) edge (\peer);

    \foreach \pos/\i in {below left of/1, below of/2}
        \node[two_hop, \pos=e3] (h\i) {};
    \foreach \speer/\peer in {e3/h1, e3/h2}
        \path (\speer) edge (\peer);

    \foreach \pos/\i in {above of/1, above right of/2}
        \node[two_hop, \pos=e4] (i\i) {};
    \foreach \speer/\peer in {e4/i1, e4/i2}
        \path (\speer) edge (\peer);
        
    \foreach \pos/\i in {above of/1, above left of/2}
        \node[two_hop, bottom color=white!30, \pos=f3] (m\i) {};    
    \foreach \speer/\peer in {f3/m1, f3/m2}
        \path (\speer) edge (\peer);

    \foreach \pos/\i in {above of/1, right of/2}
        \node[two_hop, bottom color=white!30, \pos=i2] (n\i) {};
    \foreach \speer/\peer in {i2/n1, i2/n2}
        \path (\speer) edge (\peer);

    \foreach \pos/\i in {left of/1, below of/2}
        \node[two_hop, bottom color=white!30, \pos=h1] (o\i) {};    
    \foreach \speer/\peer in {h1/o1, h1/o2}
        \path (\speer) edge (\peer);
    
    \foreach \pos/\i in {right of/1, below right of/2}
        \node[two_hop, bottom color=white!30, \pos=g1] (p\i) {};    
    \foreach \speer/\peer in {g1/p1, g1/p2}
        \path (\speer) edge (\peer);

    \foreach \pos/\i in {left of/1}
        \node[two_hop, bottom color=white!30, \pos=f2] (q\i) {};    
    \foreach \speer/\peer in {f2/q1}
        \path (\speer) edge (\peer);

\end{scope}

\node[draw, fill=orange!10, dotted, rectangle, minimum width=7.5cm, minimum height=3.5cm, rounded corners=1ex] at (5.75, 3.5) {};

\draw [decorate,decoration={brace,amplitude=5pt,raise=4pt},yshift=0pt] (10.25,2.25) -- (9.5,1.5) node [black,midway,xshift=-0.15cm, yshift=-0.75cm, rotate=45] {$N$ experts};


\node[textblock, minimum width=1cm, fill=gray!20] (ni) at (2,-1.5) {$u$};
\node[textblock, minimum width=1cm, fill=gray!20] (nj) at (4.5,-1.5) {$\{v\}$};
\node[textblock, minimum width=1cm, fill=gray!20] (eij) at (7,-1.5) {$\{e_{uv}\}$};

\node[textblock, minimum width=1cm, fill=gray!20] (hi) at (2,-0.5) {$h_u$};
\node[textblock, fill=cyan!20] (linear1) at (6,7) {FFN of $k$-th expert};
\node[textblock, minimum width=1cm, fill=green!20] (tildehi) at (6,8) {$h_{u,i\ast}$};

\node[textblock, minimum width=1.25cm, fill=lavander!60] (router) at (0,-0.5) {$router$};
\node[][draw, fill=orange!10, thick,minimum width=0.2cm,minimum height=0.5cm] () at (-0.2, 0){};
\node[][draw, fill=green, thick,minimum width=0.2cm,minimum height=0.75cm] () at (0, 0.125){};
\node[][draw, fill=orange!30, thick,minimum width=0.2cm,minimum height=0.25cm] () at (0.2, -0.125){};

\node[textblock, minimum width=1cm, fill=gray!20] (hj) at (4.5,-0.5) {$h_v$};
\node[textblock, minimum width=1cm, fill=gray!20] (ehij) at (7,-0.5) {$e^h_{uv}$};

\draw[->] (ni) -- (hi);
\draw[->] (hi) -- (router);
\draw[->] (nj) -- (hj);
\draw[->] (eij) -- (ehij);

\draw[->] (hi) -- (2,0) -- (3,1) -- (3, 1.25);
\draw[->] (2.75, 0.75) -- (2.75, 1);
\draw[->] (2.5, 0.5) -- (2.5, 0.75);

\draw[->] (hj) -- (4.5,0) -- (5.5,1) -- (5.5, 1.25);
\draw[->] (5.25, 0.75) -- (5.25, 1);
\draw[->] (5, 0.5) -- (5, 0.75);

\draw[->] (ehij) -- (7,0) -- (8,1) -- (8, 1.25);
\draw[->] (7.75, 0.75) -- (7.75, 1);
\draw[->] (7.5, 0.5) -- (7.5, 0.75);

\draw[->, dotted, thick] (0,1) -- (0,5) -- (1,6) -- (2,6);

\draw[->] (6,6.5) -- (linear1);
\draw[->] (linear1) -- (tildehi);

\node[align=left] at (3.75, 4) {\textsc{Graph Shell \#1}\\\\as processed by\\the first expert $E_1$};
\node[] at (4, 6) {\textsc{Graph Shell \#k}};

\end{scope}

\begin{scope}[transform canvas={scale=0.8}, yshift=4.75cm, xshift=12.5cm, node distance=0.6cm, every node/.append style={scale=0.8}]
    \filldraw[fill=white!10!white,thick,dashed] (-3, -0.5) circle (1.6cm);    \filldraw[fill=white!10!white,thick,dashed] (-3, -0.5) circle (1.15cm);
    \filldraw[fill=lavander!20!white,thick,dashed] (-3, -0.5) circle (0.7cm);
    \filldraw[fill=white!60!black,thick,dashed] (-3, -0.5) circle (0.25cm);

    \foreach \place/\name in {{(-3,-0.5)/e}}
        \node[superpeers] (\name) at \place {};
   %
    \foreach \pos/\i in {above left of/1, below right of/2, below left of/3, above right of/4}
        \node[one_hop, \pos=e] (e\i) {};

    \foreach \speer/\peer in {e/e1,e/e2,e/e3,e/e4}
        \path (\speer) edge (\peer);

    \foreach \pos/\i in {above left of/1, left of/2, above of/3}
        \node[two_hop, bottom color=white!30, \pos=e1] (f\i) {};
    \foreach \speer/\peer in {e1/f1, e1/f2, e1/f3}
        \path (\speer) edge (\peer);

    \foreach \pos/\i in {right of/1, below of/2}
        \node[two_hop, bottom color=white!30, \pos=e2] (g\i) {};
    \foreach \speer/\peer in {e2/g1, e2/g2}
        \path (\speer) edge (\peer);

    \foreach \pos/\i in {below left of/1, below of/2}
        \node[two_hop, bottom color=white!30, \pos=e3] (h\i) {};
    \foreach \speer/\peer in {e3/h1, e3/h2}
        \path (\speer) edge (\peer);

    \foreach \pos/\i in {above of/1, above right of/2}
        \node[two_hop, bottom color=white!30, \pos=e4] (i\i) {};
    \foreach \speer/\peer in {e4/i1, e4/i2}
        \path (\speer) edge (\peer);
        
    \foreach \pos/\i in {above of/1, above left of/2}
        \node[two_hop, bottom color=white!30, \pos=f3] (m\i) {};    
    \foreach \speer/\peer in {f3/m1, f3/m2}
        \path (\speer) edge (\peer);

    \foreach \pos/\i in {above of/1, right of/2}
        \node[two_hop, bottom color=white!30, \pos=i2] (n\i) {};
    \foreach \speer/\peer in {i2/n1, i2/n2}
        \path (\speer) edge (\peer);

    \foreach \pos/\i in {left of/1, below of/2}
        \node[two_hop, bottom color=white!30, \pos=h1] (o\i) {};    
    \foreach \speer/\peer in {h1/o1, h1/o2}
        \path (\speer) edge (\peer);
    
    \foreach \pos/\i in {right of/1, below right of/2}
        \node[two_hop, bottom color=white!30, \pos=g1] (p\i) {};    
    \foreach \speer/\peer in {g1/p1, g1/p2}
        \path (\speer) edge (\peer);

    \foreach \pos/\i in {left of/1}
        \node[two_hop, bottom color=white!30, \pos=f2] (q\i) {};    
    \foreach \speer/\peer in {f2/q1}
        \path (\speer) edge (\peer);

\end{scope}

\end{tikzpicture}
}
\caption{Routing mechanism to $N$ experts}
\label{fig:sea_routingShells}
\end{figure}
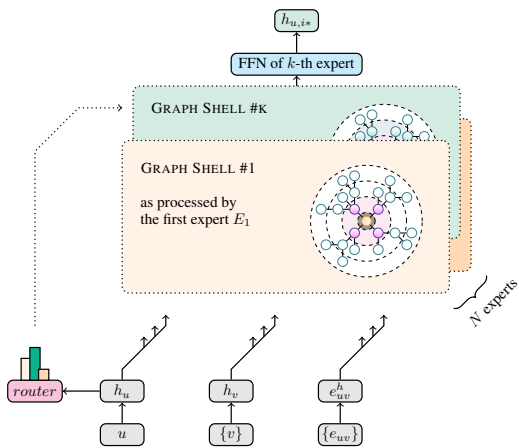

Originally introduced for language modeling and machine translation, Shazeer et al. \cite{shazeer_2017_moelayer} proposed a Mixture-of-Experts (MoE) layer. The general idea relies on a routing mechanism for token representations $x$ to determine the best expert from a set $\{E_i(x)\}_{i=1}^N$ of $N$ experts. The router module consists of a single linear transformation whose output is normalized via softmaxing over the available $N$ experts. Hence, the probability of choosing the $i$-th expert for node $u$ is given as:
\begin{equation}
    p_i(u) = \frac{\exp(r(u)_i)}{\sum_{j}^N \exp(r(u)_j)}, \quad     r(u) = h_u^T W_r
\end{equation}

\noindent%
where $r(\cdot)$ denotes the routing operation for a node $u$ with $W_r \in \mathbb{R}^{d\times N}$ being the routing's learnable weight matrix.
The idea is to select the expert $E_{i\ast}(\cdot)$ that is the most representative for a node's representation, i.e, where $i\ast=\underset{i=1,\ldots, N}{\argmax}~p_i(u)$. A node's representation calculated by the chosen expert is then used as input for an expert's individual linear transformation:
\begin{equation}
    h_{u,i\ast} =  E_{i\ast}(u)^T W_{i\ast} + b_{i\ast},
\end{equation}
\noindent%
where $W_{i\ast} \in \mathbb{R}^{d \times d}$ denotes the weight matrix of expert $E_{i\ast}(\cdot)$, $b_{i\ast}$ denotes the bias term. The node's representation according to expert $E_{i\ast}(\cdot)$, is denoted by $h_{u,i\ast}$.
The architecture on how the routing mechanism is integrated into our model is shown in figure \ref{fig:sea_routingShells}.




\definecolor{gold}{RGB}{212, 175, 55}
\definecolor{silver}{RGB}{170, 169, 173} 
\definecolor{bronze}{RGB}{176, 141, 87} 
\definecolor{lime}{RGB}{0, 255, 0}
\definecolor{navy}{RGB}{32, 33, 79}
\definecolor{graytone}{RGB}{6,6,6}

\newcolumntype{Y}{>{\centering\arraybackslash}p{2cm}}
\newcolumntype{Z}{>{\centering\arraybackslash}p{1.5cm}}
\newcolumntype{X}{>{\centering\arraybackslash}p{1cm}}
\newcolumntype{H}{>{\setbox0=\hbox\bgroup}c<{\egroup}@{}}

\newcommand{\gold}[1]{\cellcolor{gold}{\bfseries #1}}
\newcommand{\silver}[1]{\cellcolor{silver}{\bfseries #1}}
\newcommand{\bronze}[1]{\cellcolor{bronze}{\bfseries #1}}
\newcommand{\lime}[1]{\cellcolor{lime}{\bfseries #1}}

\newcommand{\blackcell}[1]{\cellcolor{graytone}{\bfseries\textcolor{white}{#1}}}

\section{Evaluation}
\label{sec:sea_evaluation}

\subsection{Experimental Setting}
\label{subsec:sea_experimentalSetting}

\subsubsection{Datasets}
\label{subsubsec:sea_datasets}\hfill\\
\textbf{ZINC} \cite{irwin_zinc_2012} is one of the most popular real-world molecular dataset consisting of 250K graphs. A subset consisting of $10$K train, $1$K validation, and $1$K test graphs is used in the literature as benchmark \cite{dwivedi_2020_benchmarking}. The task is to regress a molecular property known as the constrained solubility. For each molecular graph, the node features are types of heavy atoms, and edge features are types of bonds between them.

We also evaluate our models on \textbf{ogbg-molhiv} \cite{hu_2020_ogb}. Each graph within the dataset represents a molecule, where nodes are atoms and edges are chemical bonds. The task is to predict the target molecular property that is cast as binary label, i.e., whether a molecule inhibits HIV replication or not. It is known that this dataset suffers from a de-correlation between validation and test set performance.

A benchmark dataset generated by the \emph{Stochastic Block Model} (SBM) \cite{abbe_2018_sbms} is \textbf{PATTERN}. The graphs within this dataset do not have explicit edge features. The task is to classify the nodes into $2$ communities. The size of this dataset encompasses 14K graphs.\hfill\\

\noindent The benchmark datasets are summarized in table \ref{tab:sea_dataset_statistics}.

\begin{table*}[]
    \centering
    \resizebox{0.65\linewidth}{!}{
    \begin{tabular}{|l|l|c|l|}
         \blackcell{Domain}& \blackcell{Dataset} & \blackcell{\#Graphs} &
         \blackcell{Task}\\
         \toprule
         \multirow{2}{*}{Chemistry: Real-world molecular graphs}
          & ZINC & 12K & Graph Regression\\
          & OGBG-MOLHIV & 41K & Graph Classification\\
         \midrule
         Mathematical Modeling: Stochastic Block Models & PATTERN & 14K & Node Classification \\
         \bottomrule
    \end{tabular}
    }
    \caption{Summary dataset statistics}
    \label{tab:sea_dataset_statistics}
\end{table*}

\subsubsection{Implementation Details}
\label{subsubsec:sea_implemtnationDetails}\hfill\\
Our implementation builds upon PyTorch \cite{paszke_2019_pytorch}, DGL \cite{wang2019dgl}, and OGB \cite{hu_2020_ogb}. We trained our models on a single GPU, an NVIDIA GeForce RTX 2080 Ti. 

\subsubsection{Model Configuration}
\label{subsbusec:sea_modelConfiguration}\hfill\\
We use the Adam optimizer \cite{kingma:adam} with an initial learning rate selected in $\{10^{-3}, 10^{-4}\}$. We apply the same learning rate decay strategy for all models that half the learning rate if the validation loss does not improve over a fixed number of $5$ epochs. 

We tune the pairing $(\# \text{heads}, \text{hidden dimension}) \in \{(4,32), (8,64), (8,56))\}$ and use $\textsc{readout} \in \{sum\}$ as function for inference on the whole graph information. \emph{Batch Normalization} and \emph{Layer Normalization} are disabled, whereas residual connections are activated per default. For dropout we tuned the value to be $\in \{0, 0.01, 0.05, 0.07, 0.1\}$ and a weight decay $\in \{0, \text{5e-5}\}$. For the number of graph shells, i.e, number of experts being used, we report values $\in \{4, 6, 8, 10,  12\}$. As aggregation functions we use $\textsc{Aggregate} \in \{sum\}$ and $\textsc{Aggregate}_\mu \in \{mean\}$. For LPE, the $8$ smallest eigenvectors are used.

\subsection{Prediction Tasks}
In the following series of experiments, we investigate the performance of the \emph{Graph Shell Attention} mechanism on graph-level prediction tasks for the datasets ogbg-molhiv \cite{hu_2020_ogb} and ZINC \cite{irwin_zinc_2012}, and a node-level classification task on PATTERN \cite{abbe_2018_sbms}. We use commonly used metrics for the prediction tasks, i.e., mean absolute error (MAE) for ZINC, the ROC-AUC score on ogbg-molhiv, and the accuracy on PATTERN. \hfill\\

\noindent\textbf{Competitors. }We evaluate our architectures against state-of-the-art GNN models achieving competitive results. Our report subsumes the vanilla GCN \cite{Kipf_gcn_2017}, GAT \cite{velickovic_gat_2017} that includes additional attention heuristics, or more recent GNN architectures building on top of Transformer-enhanced models like SAN \cite{kreuzer_2021_rethinking} and Graphormer \cite{ying2021_graphormer}. Moreover, we include GIN \cite{xu_2018_howpowerfulgnn} that is more discriminative towards graph structures compared to GCN \cite{Kipf_gcn_2017}, GraphSage \cite{Hamilton_GraphSage_2017}, and DGN \cite{Beaini_dgn_2020} being more discriminative than standard GNNs w.r.t the Weisfeiler-Lehman 1-WL test. \hfill\\

\begin{table*}[!h]
    \centering
    \begin{minipage}[t]{1\columnwidth}
        \resizebox{1\textwidth}{!}{
\begin{tabular}{|c| Z|Y|}
        \toprule
        ~& \multicolumn{2}{c|}{\bfseries ZINC}\\
        \bottomrule
        \blackcell{Model} & \blackcell{\#params.} & \blackcell{MAE} \\
        \toprule
        GCN~\cite{Kipf_gcn_2017} & 505K & 0.367\\
        GIN~\cite{Xu_gin_2018} & 509K & 0.526\\
        GAT~\cite{velickovic_gat_2017} & 531K & 0.384\\
        SAN~\cite{kreuzer_2021_rethinking}& 508K & \silver{0.139}\\
        Graphormer-{\small \textsc{Slim}}~\cite{ying2021_graphormer} & 489K & \gold{0.122} \\
\midrule
        Vanilla GTL & 83K & 0.227 \\
        SEA-GNN & 347K & 0.212\\
        SEA-\textsc{aggregated} & 112K & 0.215\\
        SEA-\textsc{$2$-hop} & 430K & \bronze{0.159}\\ SEA-\textsc{$2$-hop-aug} & 709K & 0.189\\       
        \bottomrule
    \end{tabular}
        }
        \caption{Comparison to state-of-the-art on ZINC \cite{irwin_zinc_2012}; results are partially taken from \cite{kreuzer_2021_rethinking, dwivedi_2020_benchmarking}; color coding (gold/silver/bronze)}
        \label{tab:sea_comparison_zinc}
    \end{minipage}
    \hfill
    \begin{minipage}[t]{1\columnwidth}
        \resizebox{1\textwidth}{!}{
\begin{tabular}{|c| Z|Y|}
        \toprule
        & \multicolumn{2}{c|}{\bfseries OGBG-MOLHIV}\\
        \bottomrule
        \blackcell{Model} & \blackcell{\#params.} & \blackcell{\%ROC-AUC}\\
        \toprule
        GCN-{\small \textsc{GraphNorm}}~\cite{Kipf_gcn_2017} & 526K & 76.06 \\
        GIN-{\small \textsc{VN}}~\cite{Xu_gin_2018} & 3.3M & 77.80 \\
        DGN~\cite{Beaini_dgn_2020} & 114K & 79.05\\
        Graphormer-{\small \textsc{Flag}}~\cite{ying2021_graphormer} & 47.0M & \gold{80.51}\\
\midrule
        Vanilla GTL & 386K & 78.06\\ 
        SEA-GNN & 347K & 79.53\\
        SEA-\textsc{aggregated} & 133K & \silver{80.18}\\
        SEA-\textsc{$2$-hop} & 511K & \bronze{80.01}\\
        SEA-\textsc{$2$-hop-aug} & 594K & 79.08\\    
        \bottomrule
    \end{tabular}

        }
        \caption{Comparison to state-of-the-art on ogbg-molhiv \cite{hu_2020_ogb}; results are partially taken from \cite{kreuzer_2021_rethinking, dwivedi_2020_benchmarking}; color coding (gold/silver/bronze)}
        \label{tab:sea_comparison_molhiv}
    \end{minipage}
\end{table*}

\begin{table}
\centering
\resizebox{1\linewidth}{!}{
\begin{tabular}{|c| Z|Y|}
        \toprule
        & \multicolumn{2}{c|}{\bfseries PATTERN}\\
        \bottomrule
        \blackcell{Model} & \blackcell{\#params.} & \blackcell{\% ACC}  \\
        \toprule
        GCN~\cite{Kipf_gcn_2017} & 500K & 71.892\\
        GIN~\cite{Xu_gin_2018} & 100K & 85.590\\
        GAT~\cite{velickovic_gat_2017} & 526K & 78.271\\
        GraphSage~\cite{Hamilton_GraphSage_2017} & 101K & 50.516 \\
        SAN~\cite{kreuzer_2021_rethinking} & 454K & \bronze{86.581}\\
\midrule
        Vanilla GTL & 82K & 84.691\\
        SEA-GNN & 132K & 85.006\\
        SEA-\textsc{aggregated} & 69K & 57.557\\
        SEA-\textsc{$2$-hop} & 48K & \gold{86.768}\\     SEA-\textsc{$2$-hop-aug} & 152K & \silver{86.673}\\    
        \bottomrule
    \end{tabular}
}
    \caption{Comparison to state-of-the-art on PATTERN; results are partially taken from \cite{kreuzer_2021_rethinking, dwivedi_2020_benchmarking}; color coding (gold/silver/bronze)};
    \label{tab:sea_comparison_pattern}
\end{table}

\noindent\textbf{Results. }Tables \ref{tab:sea_comparison_zinc}, \ref{tab:sea_comparison_molhiv}, and \ref{tab:sea_comparison_pattern} summarize the performances of our SEA models compared to baselines on ZINC, ogbg-molhiv, and PATTERN. 
\emph{Vanilla GTL} shows the results of our implementation of the GNN model including Graph Transformer Layers \cite{dwivedi_2021_generalization}. 
\emph{SEA-\textsc{2-hop}} includes the 2-hop connection within the input graph, whereas \emph{SEA-\textsc{2-hop-aug}} process the input data the same way as the \emph{\textsc{2-hop}} heuristic, but uses additional feedforward networks for computing $Q$, $K$, $V$ values for the $2$-hop neighbors.

For PATTERN, we observe the best result using the \emph{SEA-\textsc{2-hop}} model, beating all other competitors. On the other hand, distributing an aggregated value to neighboring nodes according to \emph{SEA-\textsc{aggregated}} yields a too coarse view for graphs following the SBM and loses local graph structure. 

In the sense of \emph{Green AI} \cite{schwartz_2020_greenai} that focuses on reducing the computational cost to encourage a reduction in the resources spent, our architecture reaches state-of-the-art performance on ogbg-molhiv while reducing the number of parameters being trained. 
Comparing \emph{SEA-\textsc{aggregated}} to the best result reported for \emph{Graphormer} \cite{ying2021_graphormer}, our model economizes on $\mathbf{99.71\%}$ of the number of parameters while still reaching competitive results.


The results on ZINC enforces the argument of using individual experts compared to vanilla GTLs, where the best result is reported for \emph{SEA-\textsc{2-hop}}.\ 

\subsection{Number of Shells}
\begin{table*}
\centering
\resizebox{0.95\linewidth}{!}{
\begin{tabular}{|c|Z||ZYX|ZYX|ZYX|}
  \toprule
    & & \multicolumn{3}{c|}{\bfseries ZINC} 
      & \multicolumn{3}{c|}{\bfseries OGBG-MOLHIV} 
      & \multicolumn{3}{c|}{\bfseries PATTERN}\\
    \blackcell{Model} & \blackcell{\#experts} 
      & \blackcell{\#params} & \blackcell{MAE} & \blackcell{time/ epoch} 
      & \blackcell{\#params.} & \blackcell{\%ROC-AUC} & \blackcell{time/ epoch} 
      & \blackcell{\#params.} & \blackcell{\% ACC} & \blackcell{time/ epoch}
    \\
    \toprule
    \multirow{2}{*}{SEA-GNN}
    & 4 & 183K & 0.385 & 13.60
        & 182K & 79.24 & 49.21
        & 48K & 78.975 & 58.14\\
    & 6 & 266K & 0.368 & 20.93
        & 263K & 78.24 & 68.67
        & 69K & 82.117 & 82.46 \\
    & 8 & 349K & \lime{0.212} & 26.24
        & 345K & \lime{79.53} & 84.35
        & 90K & 82.983 & 108.41\\
    & 10 & 433K & 0.264 & 31.63
        & 428K & 79.35 & 107.11
        & 111K & 84.041 & 133.73\\
    & 12 & 516K & 0.249 & 38.26
        & 511K & 79.18 & 122.99
        & 132K & \lime{85.006} & 168.47\\
    \midrule
    \multirow{2}{*}{SEA-\textsc{aggregated}}
    & 4 & 49K & 0.257 & 31.24
        & 48K & 77.87 & 60.98
        & 48K & 57.490 & 99.10\\
    & 6 & 70K & 0.308 & 44.61
        & 69K & 79.21 & 86.26
        & 69K & \lime{57.557} & 106.79\\
    & 8 & 91K & 0.249 & 57.89
        & 90K & 77.19 & 86.93 
        & 90K & 54.385 & 131.57\\
    & 10 & 112K & \lime{0.215} & 73.49
        & 111K & 77.48 & 102.40 
        & 111K & 57.221 & 173.74\\
    & 12 & 133K & 0.225 & 87.08
        & 132K & \lime{80.18} & 124.08
        & 132K & 57.270 & 206.73\\
    \midrule
    \multirow{2}{*}{SEA-\textsc{$2$-hop}}
    & 4 & 182K & 0.309 & 14.28
        & 180K & 76.30 & 43.51
        & 48K & \lime{86.768} & 94.04\\
    & 6 & 265K & 0.213 & 20.13
        & 263K & 77.27 & 59.82
        & 69K & 86.706 & 138.10\\
    & 8 & 347K & 0.185 & 24.91
        & 345K & 76.61 & 79.56
        & 90K & 86.707 & 178.64 \\
    & 10 & 430K & \lime{0.159} & 32.68
        & 428K & 78.38 & 95.69
        & 111K & 86.680 & 232.91\\
    & 12 & 513K & 0.188 & 38.73
        & 511K & \lime{80.01} & 112.93
        & 132K & 86.699 & 269.71\\
    \midrule
    \multirow{2}{*}{SEA-\textsc{$2$-hop-aug}}
    & 4 & 248K & 0.444 & 16.86
        & 248K & 77.21 & 48.65
        & 65K & 84.889 & 124.96\\
    & 6 & 363K & 0.350 & 24.84
        & 363K & 75.19 & 70.05
        & 94K & 85.141 & 203.38\\
    & 8 & 478K & 0.285 & 31.48
        & 476K & 76.55 & 90.78 
        & 123K & 86.660 & 270.85\\
    & 10 & 594K & 0.205 & 39.25
        & 594K & \lime{79.08} & 109.91
        & 152K & \lime{86.673} & 363.58\\
    & 12 & 709K & \lime{0.189} & 46.51
        & 707K & 77.52 & 133.48
        & 181K & 86.614 & 421.46\\ 
    \bottomrule
    \end{tabular}
}
    \caption{Influence of the number of experts applied on various SEA models; best configurations are highlighted in green};
    \label{tab:sea_comparison_noexperts}
\end{table*}

Next, we examine the performance w.r.t the number of experts. Notably, increasing the number of experts correlated with the number of \emph{Graph Shells} which are taken into account. Table \ref{tab:sea_comparison_noexperts} summarizes the results where all other hyperparameters are frozen, and we only have a variable size in the number of experts. We train each model for $500$ epochs and report the best-observed metrics on the test datasets. We apply an early stopping heuristic, where we stop the learning procedure if we have not observed any improvements w.r.t the evaluation metrics or if the learning rate scheduler reaches a minimal value which we set to $10^{-6}$. Each evaluation on the test data is conducted after $5$ epochs, and the early stopping is effective after $10$ consecutive evaluations on the test data with no improvements. First, note that increasing the number of experts also increases the model's parameters linearly. This is due to additional routings and linear layer being defined for each expert separately. Secondly, we report also the average running time in seconds [$s$] on the training data for each epoch. By construction, the running time correlates with the number of parameters that have to be trained. The number of parameters differs from one dataset to another with the same settings due to a different number of nodes and edges within the datasets and slightly differs if biases are used or not. Note that we observe better results of \emph{SEA-\textsc{aggregated}} by decreasing the embedding size from $64$ to $32$, which also applies for the PATTERN dataset in general. The increase of parameters of the augmented $2$-hop architecture \emph{SEA-\textsc{2-hop-aug}} is due to the additional feedforward layers being used for the $k$-hop neighbors to compute the inputs $Q, K, V$ in the graph transformer layer. 
Notably, we also observe that similar settings apply for datasets where the structure is an important feature of the graph, like in molecules (ZINC + ogbg-molhiv). In contrast to that is the behavior on graphs following the stochastic block model (PATTERN). On the latter one, the best performance could be observed by including $k$-hop information, whereas an aggregation yields too simplified features to be competitive. For the real-world molecules datasets, we observe a tendency that more experts boost the performance.

\subsection{Stretching Locality in SEA-\textsc{k-hop}}
\begin{table}
\centering
\resizebox{1\linewidth}{!}{
\begin{tabular}{|c|c|c||cc|cZ|cc|}
  \toprule
    &&& \multicolumn{2}{c|}{\bfseries ZINC} 
      & \multicolumn{2}{c|}{\bfseries OGBG-MOLHIV} 
      & \multicolumn{2}{c|}{\bfseries PATTERN}\\
    \blackcell{Model} & \blackcell{\#exp.} & \blackcell{$k$} 
      & \blackcell{\#prms} & \blackcell{MAE}
      & \blackcell{\#prms} & \blackcell{\%ROC-~AUC}
      & \blackcell{\#prms} & \blackcell{\%ACC}
    \\
    \toprule
    \multirow{6}{*}{SEA-\textsc{k-hop}} &
    \multirow{3}{*}{6}
    & 2 & 265K & 0.213 
        & 263K & \lime{77.27}
        & 69K & \lime{86.768} \\
    &&3 & 266K & \lime{0.191} 
        & 263K & 76.15
        & 69K & 86.728 \\
    &&4 & 266K & 0.316
        & 263K & 73.48
        & 69K & 86.727\\
    \cmidrule{2-9}
    &
    \multirow{3}{*}{10}
    & 2 & 430K & \lime{0.159}
        & 428K & \lime{78.38} 
        & 111K & 86.680 \\
    &&3 & 433K & 0.171
        & 428K & 74.67
        & 111K & \lime{86.765} \\
    &&4 & 433K & 0.239
        & 428K & 73.72
        & 111K & 86.725\\    \midrule
    \end{tabular}
}
    \caption{Influence of parameter $k$ for the \emph{SEA-\textsc{k-hop}} model; best configuration for each model is highlighted in green};
    \label{tab:sea_comparison_khop}
\end{table}

In the last series of experiments, we investigate the influence of the parameter $k$ for the \emph{SEA-\textsc{k-hop}} model. Generally, by increasing the parameter $k$, the model diverges to the full model being also examined for the \emph{SAN} architecture explained in \cite{kreuzer_2021_rethinking}. In short, the full setting takes edges into account that is given by the input data and also sends information over non-existent edges, i.e., the argumentation is on a full graph setting. In our model, we smooth the transition from edges being given in the input data to the full setting that naturally arises when $k$, the number of hops, is set to a sufficiently high number. 
The table \ref{tab:sea_comparison_khop} summarizes the results for the non-augmented model, i.e., no extra linear layers are used for each $k$-hop neighborhood. The number of parameters stays the same by increasing $k$.  

\section{Conclusion \& Outlook}
\label{sec:sea_conclusion}
We introduced the theoretical foundation for integrating an expert heuristic within transformer-based graph neural networks (GNNs). This opens a fruitful direction for future works that go beyond successive neighborhood aggregation (message-passing) to develop even more powerful architectures in graph learning.

We provide an engineered solution that allows selecting the most representative experts for nodes in the input graph. For that, our model exploits the idea of a routing layer, which enables to steer the nodes' representations towards the individual expressiveness of dedicated experts.

As experts process different subgraphs starting from a central node, we introduce the terminology of \emph{Graph \textbf{S}h\textbf{e}ll \textbf{A}ttention} (SEA), where experts solely process nodes that are in their respective receptive field. Therefore, we gain expressiveness by capturing varying short- and long-term dependencies expressed by individual experts.

In a thorough experimental study, we show on real-world benchmark datasets that the gained expressiveness yields competitive performance compared to state-of-the-art results while reducing the number of parameters. Additionally, we report a series of experiments that stress the number of graph shells that are taken into account.

In the future, we aim to work on more novel implementations and applications enhanced with the graph shell attention mechanism, e.g., where different hyperparameters are used for different graph shells.

\bibliography{aaai22}

\appendix
\hfill\\
\begin{table*}[h]
\centering
\resizebox{1\linewidth}{!}{
\begin{tabular}{|c||ZZZZ|ZZZZ|ZZZZ|}
  \toprule
      & \multicolumn{4}{c|}{\bfseries ZINC} 
      & \multicolumn{4}{c|}{\bfseries OGBG-MOLHIV} 
      & \multicolumn{4}{c|}{\bfseries PATTERN}\\
    \blackcell{Param.} 
      & \blackcell{SEA-\textsc{gnn}} & \blackcell{SEA-\textsc{agg}} & \blackcell{SEA-\textsc{$2$-hop}} &
      \blackcell{SEA-\textsc{$2$-hop-aug}}
      & \blackcell{SEA-\textsc{gnn}} & \blackcell{SEA-\textsc{agg}} & \blackcell{SEA-\textsc{$2$-hop}} &
      \blackcell{SEA-\textsc{$2$-hop-aug}}
      & \blackcell{SEA-\textsc{gnn}} & \blackcell{SEA-\textsc{agg}} & \blackcell{SEA-\textsc{$2$-hop}} &
      \blackcell{SEA-\textsc{$2$-hop-aug}}
    \\
    \toprule
    batch size 
      & 64 & 32 & 32 & 64 
      & 64 & 64 & 64 & 64
      & 16 & 16 & 16 & 16 \\
    \midrule
    \midrule
    optimizer
      &\multicolumn{12}{c|}{Adam}\\
    \cmidrule{2-13}
    learning rate (lr)
      &\multicolumn{4}{c|}{$10^{-3}$} 
      &\multicolumn{4}{c|}{$10^{-3}$} 
      &\multicolumn{4}{c|}{$10^{-4}$} \\
    \cmidrule{2-13} 
    lr reduce factor & \multicolumn{12}{c|}{0.5}\\
    \cmidrule{2-13} 
    momentum &\multicolumn{12}{c|}{0.9}\\
    \midrule
    \midrule
    emb.size
      & 64 & 32 & 64 & 64
      & 64 & 32 & 64 & 64 
      & 32 & 32 & 32 & 32\\
    \cmidrule{2-13} 
    \# heads
      & 8 & 4 & 8 & 8 
      &\multicolumn{4}{c|}{4}
      &\multicolumn{4}{c|}{4} \\
    \cmidrule{2-13} 
    \# experts
      & 8 & 10 & 10 & 12 
      & 8 & 12 & 12 & 10 
      & 12 & 6 & 4 & 10\\
    \cmidrule{2-13}
    dropout
      & 0.07 & 0.01 & 0.07 & 0.07 
      & \multicolumn{4}{c|}{0.01}
      & \multicolumn{4}{c|}{0.00}\\
    \cmidrule{2-13}    
    \textsc{readout}& \multicolumn{12}{c|}{sum}\\
    \cmidrule{2-13}
    layer norm. & \multicolumn{12}{c|}{false}\\
    \cmidrule{2-13}
    batch norm. & \multicolumn{12}{c|}{false}\\
    \cmidrule{2-13}
    residual & \multicolumn{12}{c|}{true}\\
    \cmidrule{2-13}
    use bias & \multicolumn{12}{c|}{false}\\
    \cmidrule{2-13}      
    use edge features & \multicolumn{12}{c|}{false}\\
    \cmidrule{2-13}      
    LPE & \multicolumn{12}{c|}{true}\\
    \cmidrule{2-13}      
    LPE dim & \multicolumn{12}{c|}{8}\\
    \bottomrule
    \end{tabular}
}
    \caption{Hyperparameter settings}
    \label{tab:sea_hyperparameters}
\end{table*}

\begin{figure}[b]
\begin{subfigure}[t]{0.95\columnwidth}
    \begin{subfigure}[t]{0.32\columnwidth}
    \centering
    \includegraphics[width=\textwidth]{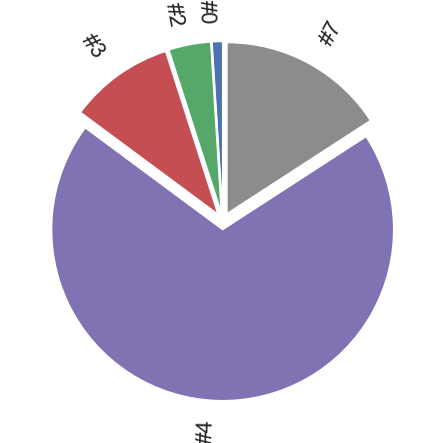}
    ZINC
    \end{subfigure}
    \begin{subfigure}[t]{0.32\columnwidth}
    \centering
    \includegraphics[width=\textwidth]{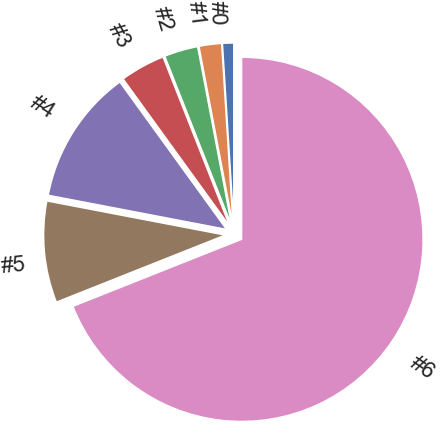}
    ogbg-molhiv
    \end{subfigure}
    \begin{subfigure}[t]{0.32\columnwidth}
    \centering
    \includegraphics[width=\textwidth]{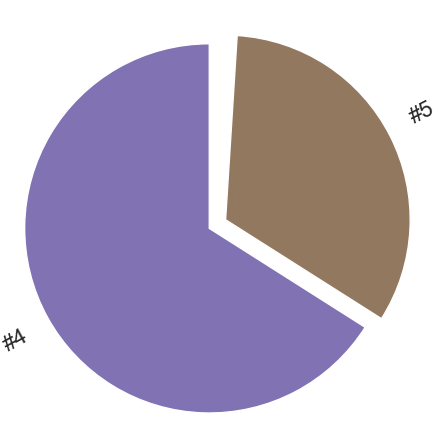}
    PATTERN
    \end{subfigure}
    \subcaption{SEA-\textsc{gnn}}
 \end{subfigure}
 
 \begin{subfigure}[t]{0.95\columnwidth}
    \begin{subfigure}[t]{0.32\columnwidth}
    \centering
    \includegraphics[width=\textwidth]{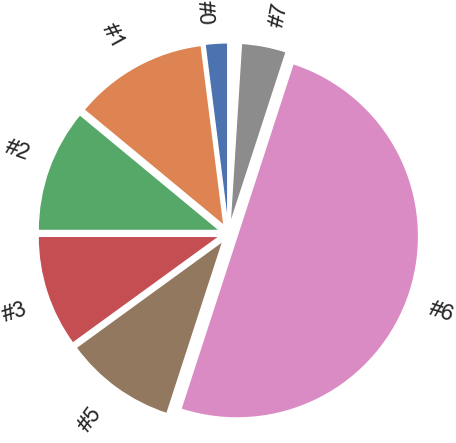}
    ZINC
    \end{subfigure}
    \begin{subfigure}[t]{0.32\columnwidth}
    \centering
    \includegraphics[width=\textwidth]{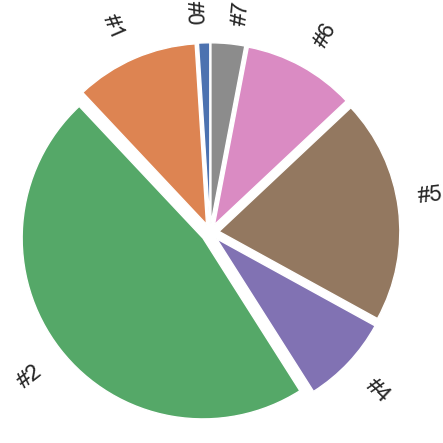}
    ogbg-molhiv
    \end{subfigure}
    \begin{subfigure}[t]{0.32\columnwidth}
    \centering
    \includegraphics[width=\textwidth]{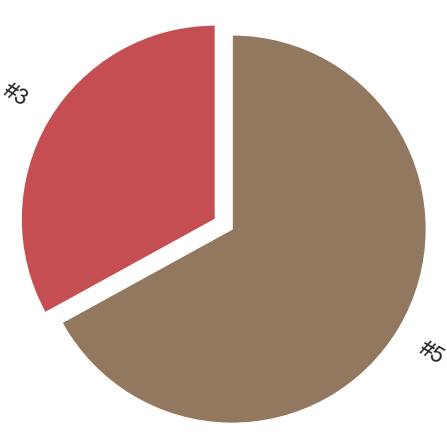}
    PATTERN
    \end{subfigure}
    \subcaption{SEA-\textsc{aggregated}}
 \end{subfigure}
 
\begin{subfigure}[t]{0.95\columnwidth}
    \begin{subfigure}[t]{0.32\columnwidth}
    \centering
    \includegraphics[width=\textwidth]{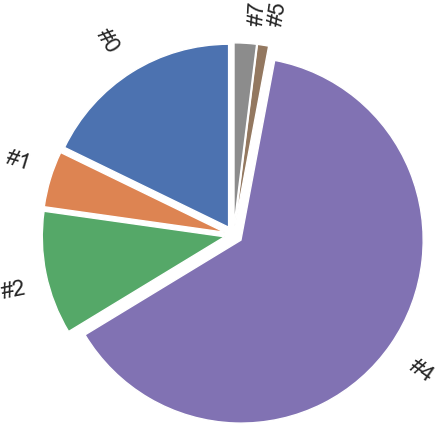}
    ZINC
    \end{subfigure}
    \begin{subfigure}[t]{0.32\columnwidth}
    \centering
    \includegraphics[width=\textwidth]{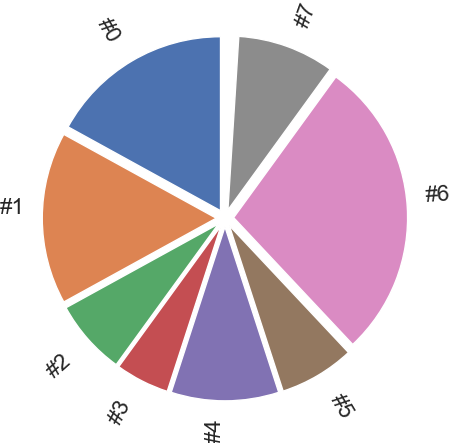}
    ogbg-molhiv
    \end{subfigure}
    \begin{subfigure}[t]{0.32\columnwidth}
    \centering
    \includegraphics[width=\textwidth]{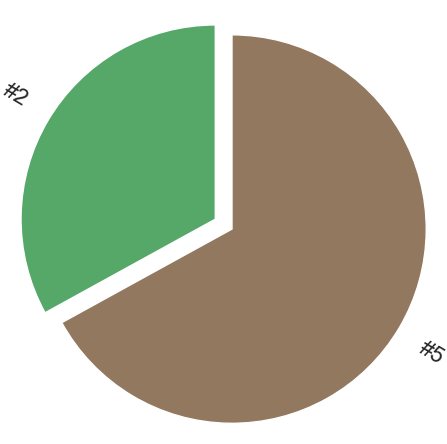}
    PATTERN
    \end{subfigure}
    \subcaption{SEA-\textsc{2-hop}}
 \end{subfigure}
 
 \begin{subfigure}[t]{0.95\columnwidth}
    \begin{subfigure}[t]{0.32\columnwidth}
    \centering
    \includegraphics[width=\textwidth]{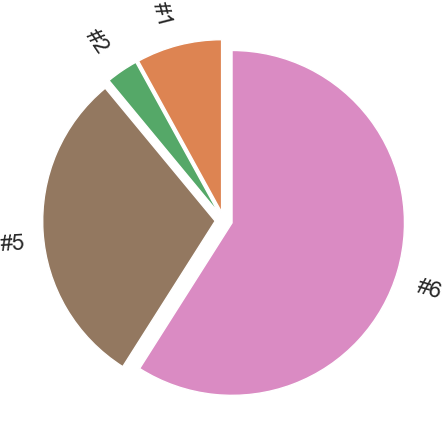}
    ZINC
    \end{subfigure}
    \begin{subfigure}[t]{0.32\columnwidth}
    \centering
    \includegraphics[width=\textwidth]{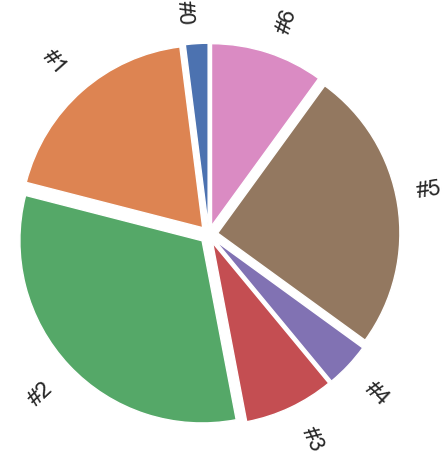}
    ogbg-molhiv
    \end{subfigure}
    \begin{subfigure}[t]{0.32\columnwidth}
    \centering
    \includegraphics[width=\textwidth]{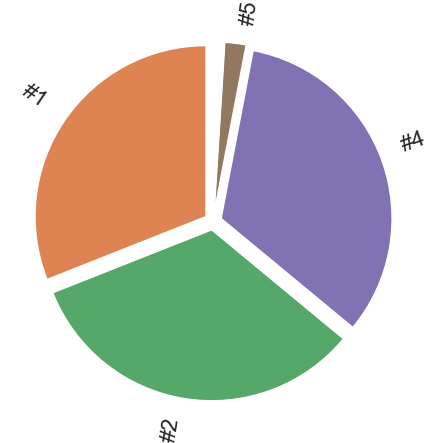}
    PATTERN
    \end{subfigure}
    \subcaption{SEA-\textsc{2-hop-aug}}
 \end{subfigure}
    \caption{Distribution of 8 experts for models \emph{SEA-\textsc{GNN}}, \emph{SEA-\textsc{aggregated}}, \emph{SEA-\textsc{2-hop}}, and \emph{SEA-\textsc{2-hop-aug}} for datasets ZINC, ogbg-molhiv and PATTERN. Relative frequencies are shown for values $\geq 1\%$. Numbers attached to the slices refer to the respective experts.}
    \label{fig:sea_expert_distribution}
\end{figure}


\section{Supplementary Material}
In the following, we provide additional information about the hyperparameters being used in our experiments. We also discuss the distribution of the experts being selected to learn the node representations of the input graph.

\subsection{Notes on Hyperparameters}
Table \ref{tab:sea_hyperparameters} summarizes the hyperparameter settings for the models SEA-\textsc{gnn}, SEA-\textsc{aggregated}, SEA-\textsc{2-hop}, and SEA-\textsc{2-hop-aug} being used individually for the datasets ZINC, ogbg-molhiv and PATTERN. 
Remarkably, in contrast to comparable architectures \cite{dwivedi_2021_generalization, kreuzer_2021_rethinking}, we could not observe improvements with normalization techniques. Generally, batch normalization during training calculates the inputs' mean and standard deviation to a layer per mini-batch and uses these statistics to perform the standardization. Note that in our architecture, each expert learns different statistics w.r.t to the information given by the graph shells they are processing, i.e., nodes within their respective fields. Hence, standardization techniques applied on mini-batches smooth out the peculiarities captured by the experts. Consequently, we disabled normalization techniques and just kept the residual connections enabled in the computational flow of graph transformer layers (GTLs). 
Smaller batch sizes for PATTERN are applied due to a limited $11$ GB GDDR6 memory used in NVIDIA's GeForce RTX 2080 Ti, which we used in our experiments. 
Notably, we also observed that disabling biases for feedforward networks yield better results. However, increasing the number of hops $k \geq 3$ within the SEA-\textsc{k-hop} model, learning also biases result in better performances. 
In contrast to experiments shown in the approach proposed by Kreuzer et al. \cite{kreuzer_2021_rethinking}, we also get better results using \emph{Adam} as optimizer instead of \emph{AdamW}.

\subsection{Distribution of Experts}
We evaluate the distributions of the experts being chosen to compute the nodes' representations in the following. We set the number of experts to $8$. Figure \ref{fig:sea_expert_distribution} summarizes the relative frequencies of the experts being chosen on the datasets ZINC, ogbg-molhiv, and PATTERN. Generally, the performance of the shell attention heuristic degenerates whenever we observe \emph{expert collapsing}. In the extreme case, just one expert expresses the mass of all nodes, and the capability to distribute learning nodes' representations over several experts is not leveraged. To overcome \emph{expert collapsing}, we can use a heuristic where in the early stages of the learning procedure, an additional epsilon parameter $\epsilon$ introduces randomness into the algorithm. Like a decaying greedy policy in Reinforcement Learning (RL), we choose a random expert with probability $\epsilon$ and choose the expert with the highest probability according to the routing layer with a probability of $1-\epsilon$. The epsilon value slowly decays over time. This ensures that all experts' expressiveness is being explored to find the best matching one w.r.t to a node $u$ and prevents getting stuck in a local optimum. The figure shows the distribution of experts that are relevant for the computation of the nodes' representations. For illustrative purposes, values below $1\%$ are omitted. 
Generally, nodes are more widely distributed over all available experts in the molecular datasets like ZINC and ogbg-molhiv for all models compared to PATTERN following a stochastic block model. Therefore, various experts are capable of capturing individual topological characteristics of molecules better than vanilla graph neural networks for which over-smoothing might potentially occur. We also observe that the mass is distributed to only a subset of the available experts for the PATTERN dataset. Hence, the specific number of iterations is more expressive for nodes within graph structures following SBM.

\end{document}